\documentclass[acmtog]{acmart}

\usepackage{booktabs} 
\usepackage{pifont}
\usepackage{tabularx}
\usepackage{multirow} 
\usepackage{xcolor}
\usepackage{float}
\usepackage{placeins}
\usepackage{xcolor}
\usepackage{graphicx} 
\usepackage{caption}  
\usepackage{cuted}    
\usepackage{amsmath}
\usepackage[ruled]{algorithm2e} 

\SetAlFnt{\small}
\SetAlCapFnt{\small}
\SetAlCapNameFnt{\small}
\SetAlCapHSkip{0pt}

\acmJournal{TOG}

\begin{document}
\title{WordCon: Word-level Typography Control in Scene Text Rendering}

\author{Wenda Shi}
\affiliation{%
 \institution{The Hong Kong Polytechnic University}
 \city{Hong Kong}
 \country{China}
}
\email{wendashi@polyu.edu.hk}

\author{Yiren Song}
\affiliation{%
 \institution{National University of Singapore}
 \city{Singapore}
 \country{Singapore}
}
\email{yiren@nus.edu.sg}

\author{Zihan Rao}
\affiliation{%
 \institution{Chongqing Univesity}
 \city{Chongqing}
 \country{China}
}
\email{20173816@cqu.edu.cn}

\author{Dengming Zhang}
\affiliation{%
 \institution{Zhejiang University}
 \city{Hangzhou}
 \country{China}
}
\email{dmz@zju.edu.cn}

\author{Jiaming Liu}
\affiliation{%
 \institution{Tiamat AI}
 \city{Shanghai}
 \country{China}
}
\email{jmliu1217@gmail.com}

\author{Xingxing Zou$^*$}
\affiliation{%
 \institution{The Hong Kong Polytechnic University}
 \city{Hong Kong}
 \country{China}
}
\email{xingxing.zou@polyu.edu.hk}


\begin{strip}
    \centering
    \includegraphics[width=\linewidth]{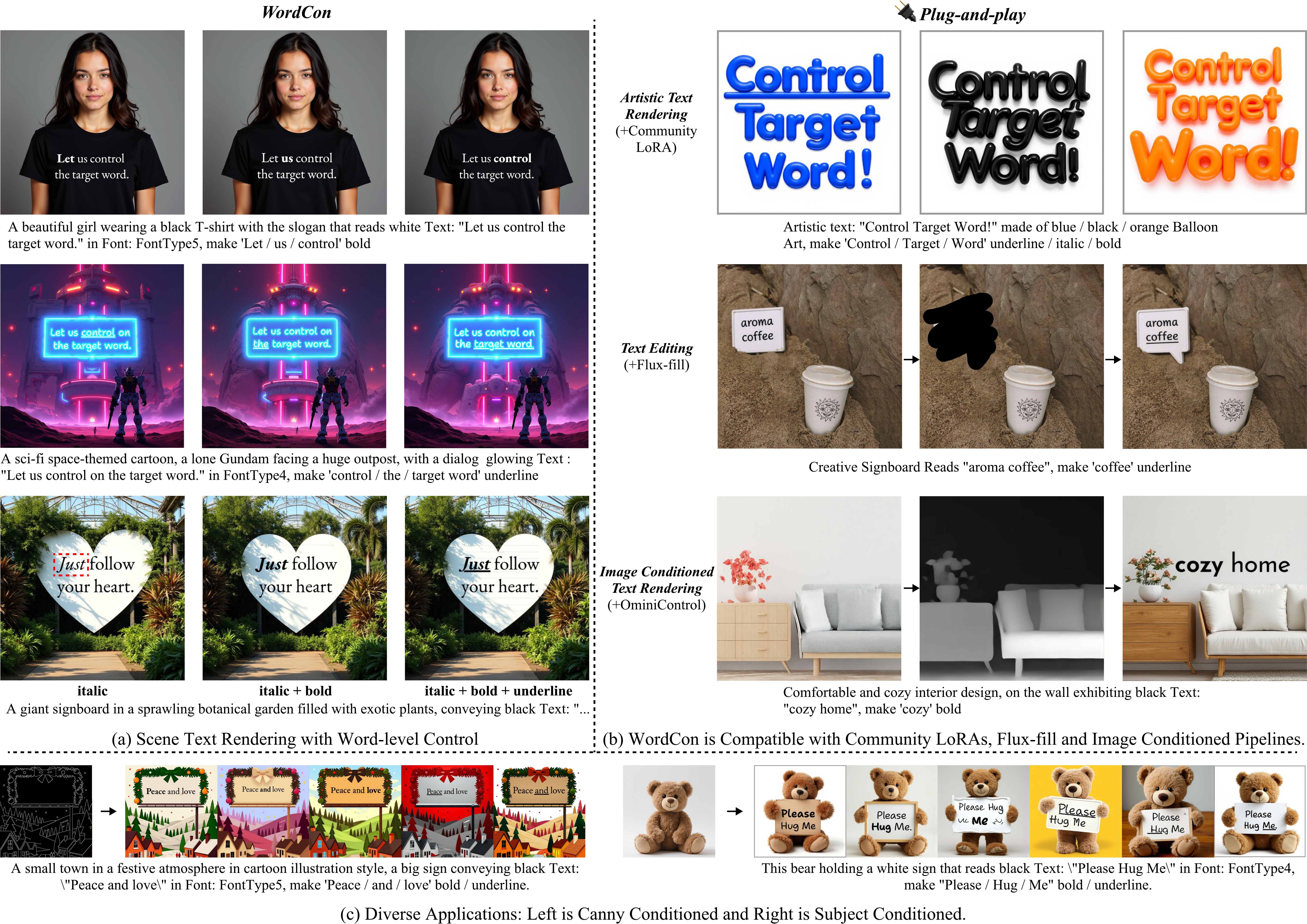}
    \vspace{-6mm}
    \captionof{figure}{(a) Scene text rendering results with word-level typography control from WordCon. The controlled content of each image is `Let', `us', `control', `the', `target word', and `Just'. The typographic attributes are `bold', `underline', and `italic' from top to bottom. (b) WordCon is compatible with artistic LoRAs, Flux-fill~\cite{FLUX-Fill}, and image conditioned pipelines~\cite{tan2024ominicontrol}, which makes it suitable for various tasks, including artistic text rendering (first row), text editing (second row), and image conditioned text rendering (third row). (c) shows more visual results of diverse applications.}
    \vspace{-5mm}
    \label{fig:teaser}
\end{strip}

\begin{abstract}
Achieving precise word-level typography control within generated images remains a persistent challenge. To address it, we newly construct a word-level controlled scene text dataset and introduce the Text-Image Alignment (TIA) framework. This framework leverages cross-modal correspondence between text and local image regions provided by grounding models to enhance the Text-to-Image (T2I) model training. Furthermore, we propose WordCon, a hybrid parameter-efficient fine-tuning (PEFT) method. WordCon reparameterizes selective key parameters, improving both efficiency and portability. This allows seamless integration into diverse pipelines, including artistic text rendering, text editing, and image-conditioned text rendering. To further enhance controllability, the masked loss at the latent level is applied to guide the model to concentrate on learning the text region in the image, and the joint-attention loss provides feature-level supervision to promote disentanglement between different words. Both qualitative and quantitative results demonstrate the superiority of our method to the state of the art. Our project is available at \url{https://wendashi.github.io/WordCon-Page/}.
\end{abstract}

%
%

%
%


\maketitle

\section{Introduction}

\begin{figure}[H]
  \centering
  \includegraphics[width=\columnwidth]{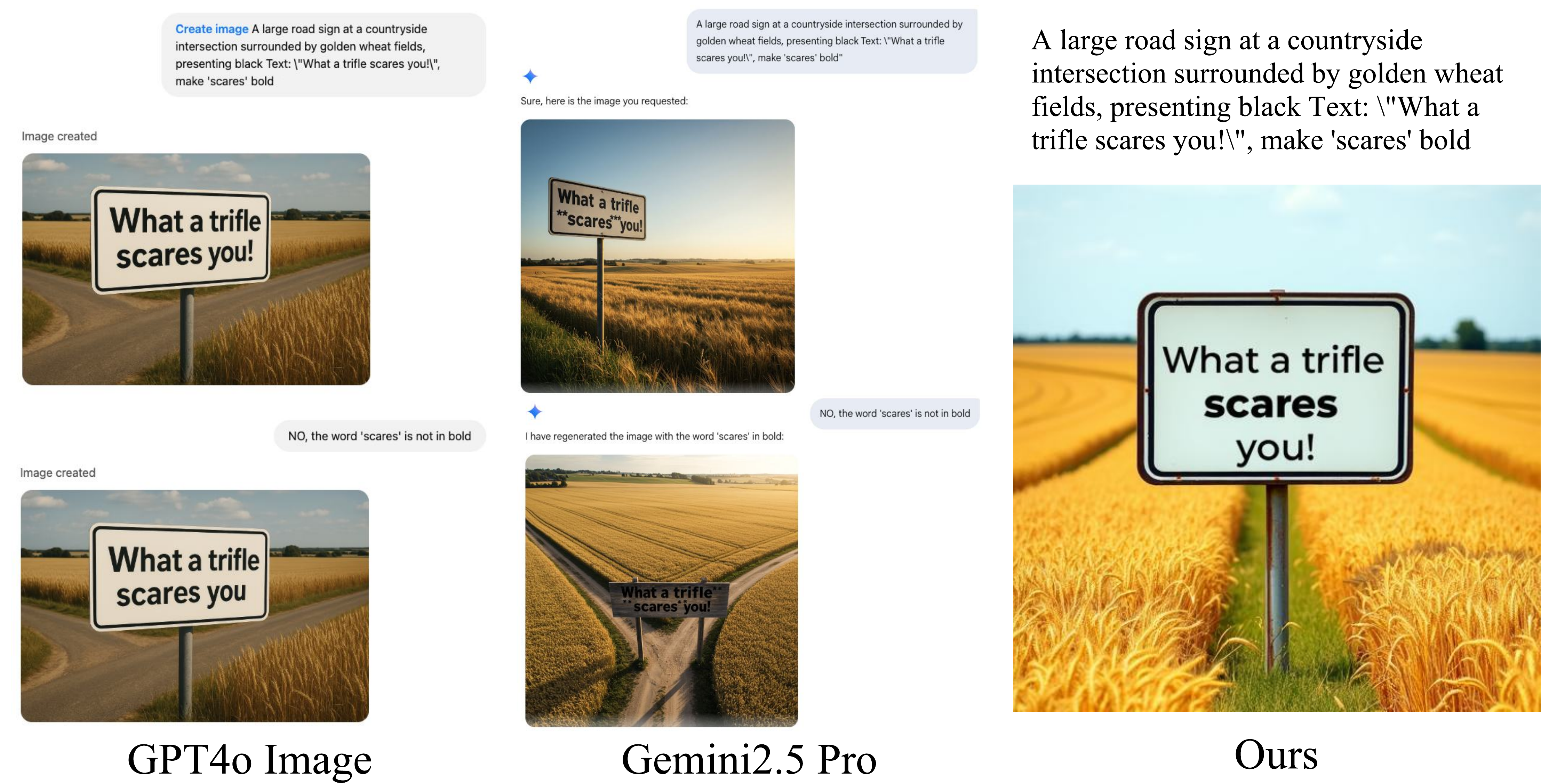}
  \vspace{-6mm}
  \caption{The challenge of text rendering. The SOTA T2I models excel at general text rendering and controllability on common objects, however, they struggle with precise word-level typography control.}
  \label{fig:problem}
  \vspace{-3mm}
\end{figure}

Scene text rendering, which involves the integration of text into images while maintaining visual coherence and realism, plays a crucial role in various applications such as advertising, branding, and marketing~\cite{bai2024intelligent, chen2024textdiffuser, zou2025fragment, shi2025generative}. However, creating high-quality visual text is a challenging task that requires designers to carefully consider font selection, typographic attributes, and overall visual harmony~\cite{shi2024fonts}.

The emergence of diffusion text-to-image (T2I) models~\cite{rombach2022high, podell2024sdxl} has shown great potential in automating this design process. However, these general-purpose models often struggle with text accuracy and controllability. To address these limitations, researchers have developed several UNet-based approaches~\cite{yang2024glyphcontrol, chen2024textdiffuser, chen2023textdiffuser, tuo2024anytext, DreamText, liu2024glyph, liu2024glyph2, zhang2025artist}. For example, Glyph-ByT5~\cite{liu2024glyph} improved text accuracy by introducing a new text encoder with contrastive learning, while TextDiffuser-2~\cite{chen2024textdiffuser} enhanced layout planning by incorporating multiple large language models.

Recent advances in diffusion transformer (DiT) models~\cite{esser2024scaling, blackforestlabs2024} greatly improved text accuracy by leveraging direct preference optimization (DPO)~\cite{rafailov2023direct} and T5~\cite{raffel2020exploring} text encoder for better prompt understanding. Additionally, closed-source models~\cite{gpt4oimg, recraft2024, ideogram2024,gimini2024} have demonstrated substantial improvements in text rendering quality and general object control. However, these models still struggle with precise word-level typography control, such as applying bold, italic, or underline to specific words, which is a capability essential for practical applications. As demonstrated in Figure~\ref{fig:problem}, even advanced commercial models like GPT4o-img~\cite{gpt4oimg} and Gemini2~\cite{gimini2024} fail to achieve the desired word-level control despite multiple attempts with various prompts.

\begin{figure}[t]
  \centering
  \includegraphics[width=\columnwidth]{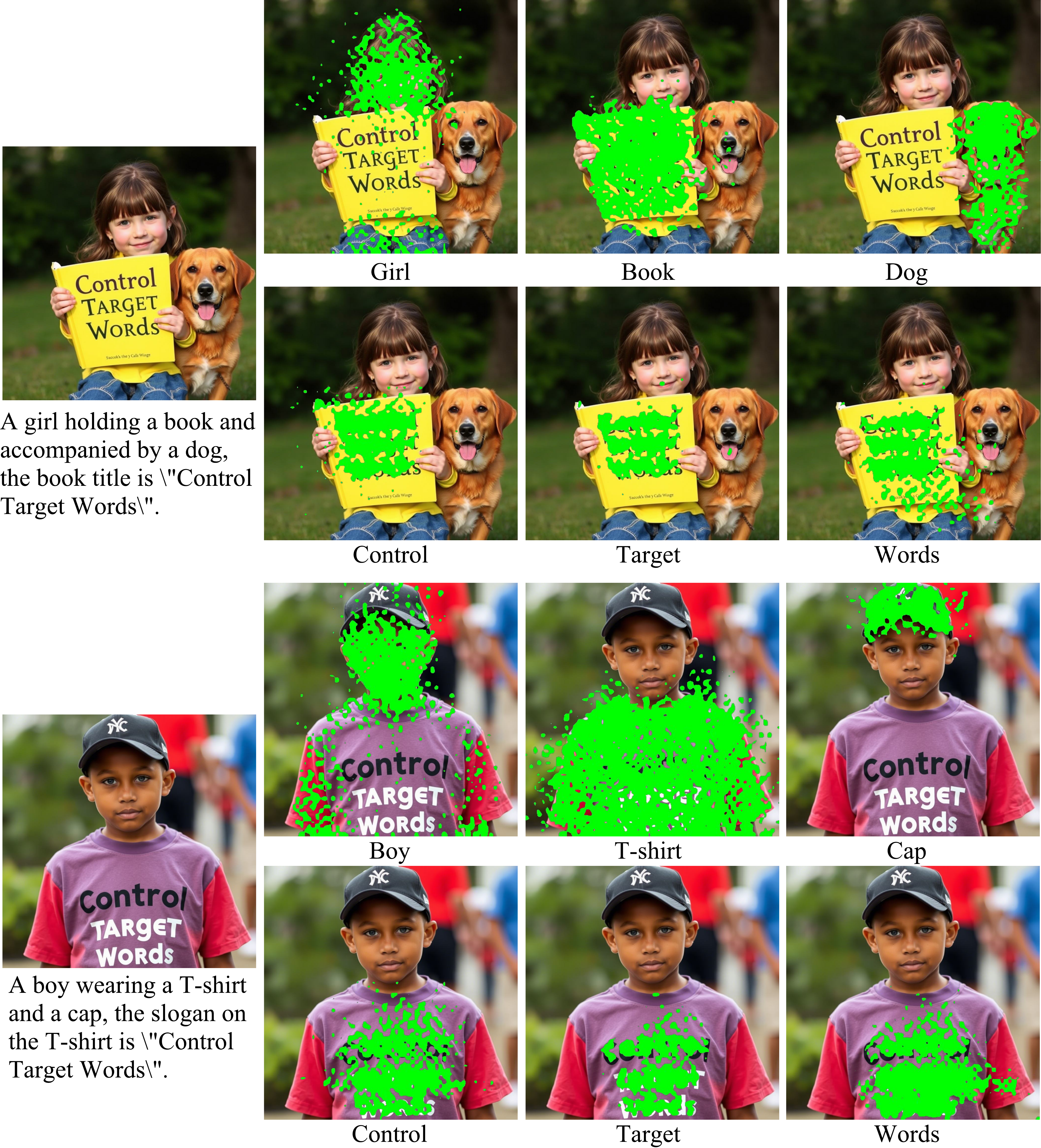}
  \vspace{-6mm}
  \caption{The green regions are the attention maps of each word in the prompt. Compared to words that refer to common objects (e.g., `Girl', `Book', `Dog', `Boy', `T-shirt', `Cap'), attention map of words for text rendering is more likely to be misaligned. In this paper, we refer to it as \textbf{word-level misalignment}.}
  \vspace{-6mm}
  \label{fig:intro_attn}
\end{figure}

To understand the reason behind this limitation, we visualized the attention maps of Flux~\cite{blackforestlabs2024} in the same analysis method as~\cite{helbling2025conceptattention, hu2025dcedit}. As shown in Figure~\ref{fig:intro_attn}, attention map misalignment occurs more prominently in text rendering than in common object generation. The decoupling between words used for text rendering is not as thorough as that between words for common objects. \textit{This finding indicates that current methods lack effective mechanisms for precise text positioning, making it challenging to achieve controllability at the word level.}

\begin{figure}[t]
  \centering
  \includegraphics[width=\columnwidth]{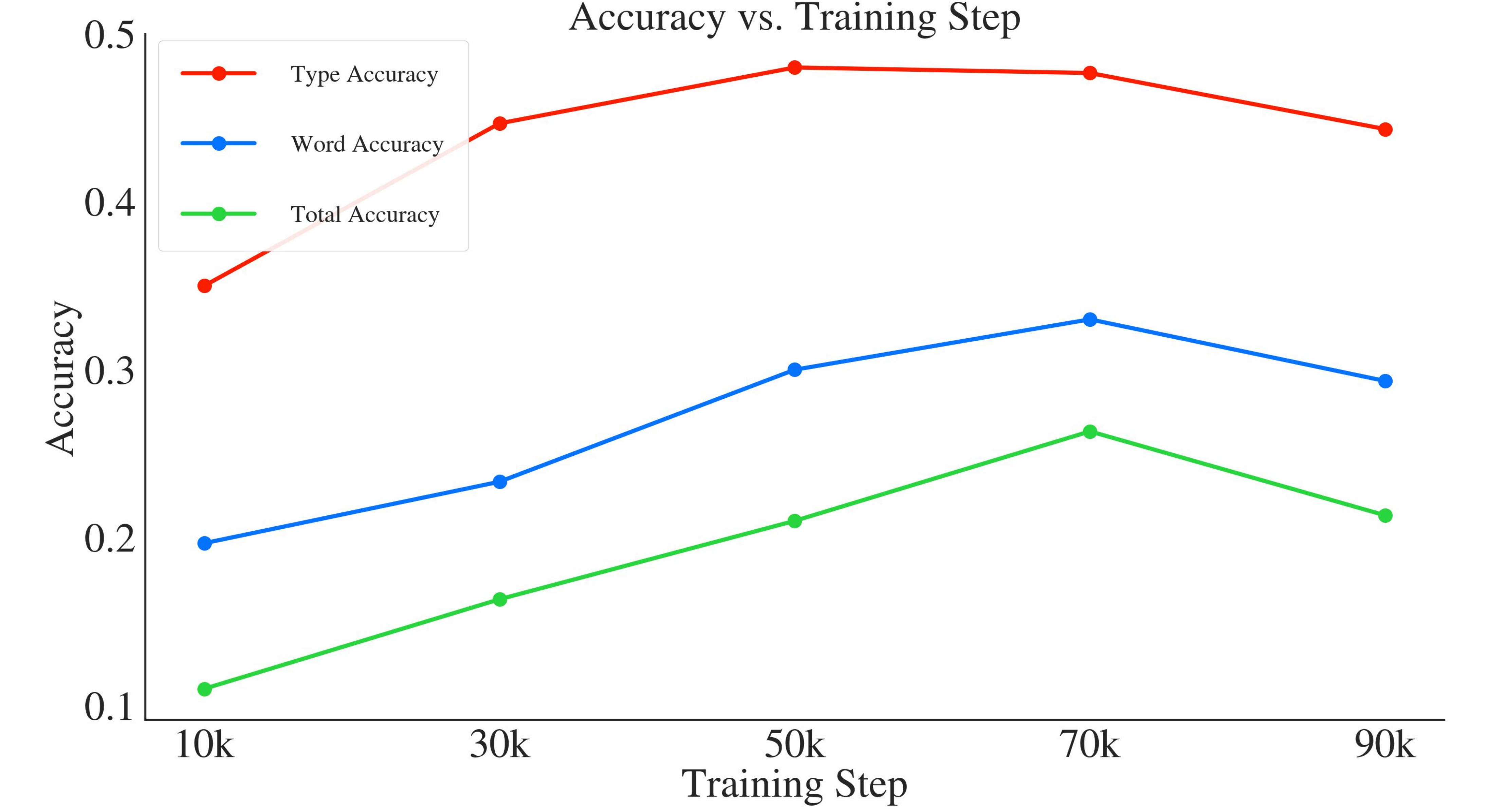}
  \vspace{-6mm}
  \caption{Results of fine-tuning Flux with word-level control dataset across varied training steps show that total acc is mainly limited by word acc.}
  \label{fig:intro_analysis}
  \vspace{-4mm}
\end{figure}

In addition, we conducted a preliminary experiment on fine-tuning Flux with a word-level control dataset. The result is shown in Figure~\ref{fig:intro_analysis}. We divide the accuracy into three types: (1) type accuracy, the typographic attribute type is correct, no matter which word is controlled; (2) word accuracy, the controlled word is correct, no matter which typographic attribute type is applied; (3) total accuracy, the attribute type and controlled word are both correct. It can be clearly seen from the curve that total accuracy is mainly limited by word accuracy. \textit{This shows that the original training goal is of limited help for the word-level misalignment problem.}

To address word-level misalignment, we propose a Text-Image Alignment (TIA) framework to leverage the fine-grained text-image alignment (text query-regional pixels) of grounding models better to align the word-level text rendering with the input prompt. Considering the growing size of DiT models, we develop WordCon, a hybrid Parameter-Efficient Fine-Tuning (PEFT) method that reduces computational costs while maintaining flexibility. With this flexibility, the module trained by our method is plug-and-play, which can be seamlessly integrated into image-conditioned pipelines~\cite{tan2024ominicontrol, zhang2025easycontrol, wu2025less}, text editing pipelines~\cite{FLUX-Fill}, and community artistic style LoRAs. Since the initial training goal is of limited help in solving the word-level misalignment, we introduce masked loss and joint-attention loss to improve the word-level controllability. Meanwhile, we construct the scene text dataset with word-level annotations and provide detailed comparative differences to facilitate model training.

Our contributions are summarized as follows:

\begin{itemize}
\item We propose TIA, a general text-image alignment framework that uses the alignment ability of grounding models to alleviate misalignment in text-to-image models.

\item We introduce WordCon, a hybrid PEFT method with supervision at the latent-level and feature-level. 

\item We construct a new scene text dataset with word-level typography control and per-word segmentation masks.

\item Comprehensive experiments with both qualitative and quantitative results demonstrate the superiority of our method.
\end{itemize}

\section{Related Work}

\textbf{Visual Text Rendering.} 
Visual text rendering can be divided into two main types: scene text rendering~\cite{yang2024glyphcontrol, chen2024textdiffuser, chen2023textdiffuser, tuo2024anytext, DreamText, liu2024glyph, liu2024glyph2, lu2025easytext} and artistic text rendering~\cite{azadi2018multi, jiang2019scfont, gao2019artistic, Yang_2019_ICCV, Mao2023IntelligentTA, tanveer2023ds, wang2023anything, mu2024fontstudio, song2022clipfont, song2023clipvg}. In \textit{scene text rendering}, despite progress in diffusion models~\cite{rombach2022high, podell2024sdxl}, high-quality scene text rendering remains a challenge. A line of research \cite{yang2024glyphcontrol, chen2024textdiffuser, chen2023textdiffuser, tuo2024anytext} focuses on explicitly controlling the position and content of the text being rendered, relying on ControlNet \cite{zhang2023adding}. e.g., methods \cite{yang2024glyphcontrol,tuo2024anytext} leverage glyph images containing multiple text lines as priors to guide diffusion models to produce accurate and accurate text, relying on ControlNet \cite{zhang2023adding}. Similarly, TextDiffuser~\cite{chen2024textdiffuser, chen2023textdiffuser} employs character-level segmentation masks as control conditions in scene text rendering. Another line of work \cite{liu2024glyph, liu2024glyph2} fine-tunes the character-aware ByT5 text encoder \cite{liu-etal-2023-character} using paired glyph-text datasets, improving the ability to render accurate text in images.

In \textit{artistic text rendering}, early research focused on font creation by transferring textures from existing characters, employing stroke-based methods~\cite{berio2022strokestyles}, patch-based techniques~\cite{yang2017awesome, yang2018context, yang2018context2}, and GAN-based~\cite{azadi2018multi, jiang2019scfont, gao2019artistic, Yang_2019_ICCV, Mao2023IntelligentTA, tang2022few} methods.  
Innovations with diffusion models~\cite{tanveer2023ds, wang2023anything, mu2024fontstudio} have enabled diverse text image stylization and semantic typography, resulting in visually appealing designs that retain readability. However, despite recent DiT models~\cite{esser2024scaling, blackforestlabs2024, shi2024fonts} showing quite promise in both artistic and scene text rendering, they still struggle with precise word-level typography control.

\noindent \textbf{PEFT of T2I Models.} 
As T2I models like Stable Diffusion series~\cite{rombach2022high, podell2024sdxl, esser2024scaling} and Flux~\cite {blackforestlabs2024} continue to grow in parameter size, Parameter-Efficient Fine-Tuning (PEFT) in T2I models has gained significant popularity. These methods can be categorized into four types: (1) \textit{Additive tuning}~\cite{ye2023ip, zhang2023adding, li2025controlnet, zhang2024ssr}, which adds trainable branches to frozen models; (2) \textit{Prompt tuning}~\cite{gal2023an}, which optimizes input prompts (new tokens for text encoder) rather than weights of backbone; (3) \textit{Selective tuning}~\cite{kumari2023multi, butt2025colorpeel}, which updates only specific key parameters of UNet backbone; and (4) \textit{Reparameterization}~\cite{tan2024ominicontrol, zhang2025easycontrol, wu2025less}, which decomposes original weights into lower-rank matrices~\cite{hu2022lora}. WordCon is a hybrid of (3) and (4) to combine parameter efficiency and portability.


\begin{figure*}[t]
  \centering
  \includegraphics[width=\textwidth]{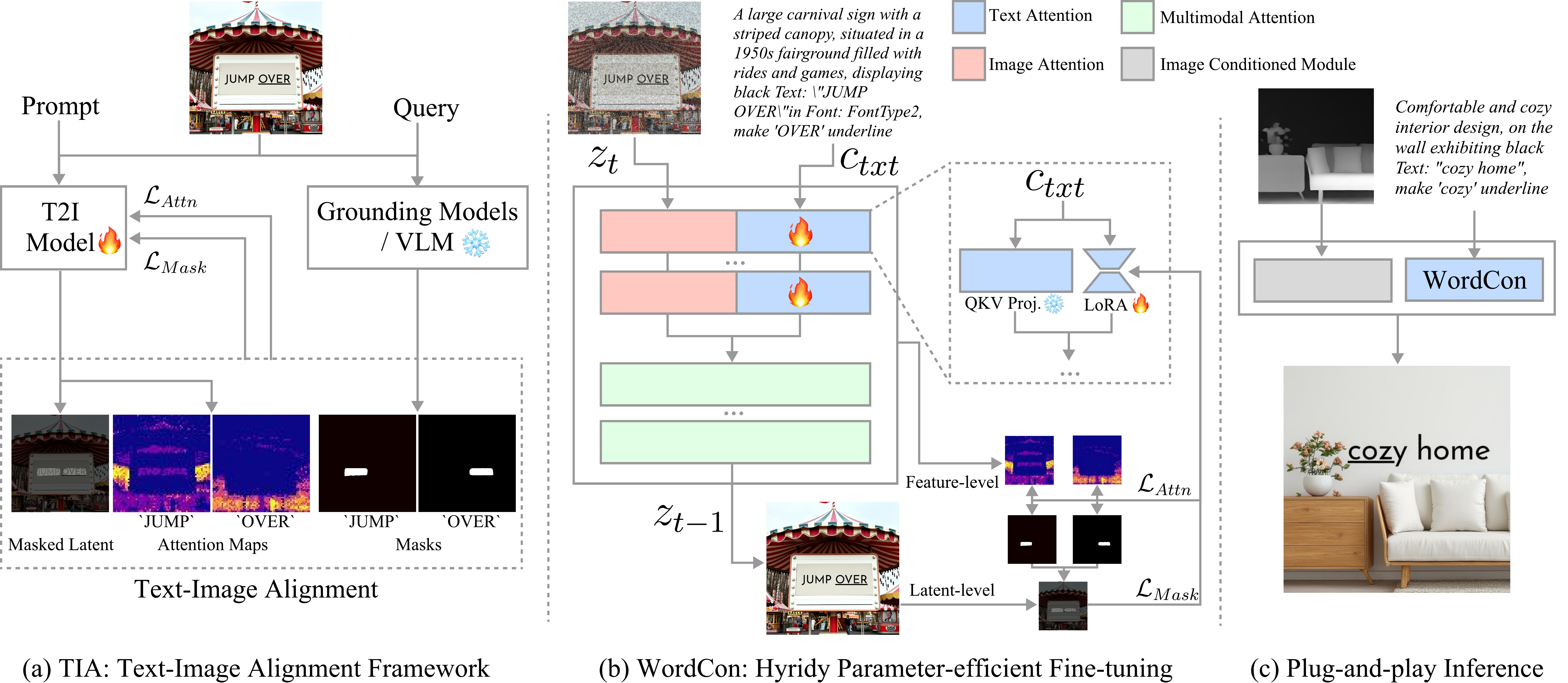}
  \vspace{-6mm}
  \caption{Method overview: (a) to mitigate word-level misalignment, we employ a text-image alignment framework that leverages the cross-modal correspondence between textual query and image regions provided by grounding models. In addition, (b) to conserve computational resources and enhance flexibility, we introduce WordCon, a hybrid PEFT method that reparameterizes selective key parameters with two losses. The masked loss at the latent level is applied to guide the model to concentrate on learning the text part, and the joint-attention loss provides feature-level supervision to promote disentanglement between different words. (c) The plug-and-play inference pipeline with other modules shows the wide applicability of our method.}
  \vspace{-3mm}
  \label{fig:method}
\end{figure*}

\noindent \textbf{Attention Alignment.} 
Attention mechanisms play a key role in T2I and visual language models (VLMs), but they have different applications in various models. \textit{In T2I models}, attention mechanisms take over the alignment between text and images and are primarily used for controlling the generation process and establishing connections between text and visual elements. HyperStyle~\cite{Alaluf2021HyperStyleSI} pioneered cross-attention maps to manipulate generated images, which was later extended to real images through inversion techniques~\cite{hertzprompt}. Attend-and-excite~\cite{chefer2023attend} further improved this approach by using cross-attention maps to enhance text-to-image alignment. CONFORM~\cite{Meral_2024_CVPR} demonstrated that these cross-attention mechanisms can produce interpretable saliency maps of textual concepts, enabling applications in object insertion~\cite{dalva2024fluxspace, avrahami2023break}, layout control~\cite{epstein2023diffusion}, and synthetic data generation~\cite{brooks2023instructpix2pix}. Despite these advances, current T2I models still struggle with precise word-level control for typography applications.

\textit{In VLMs}, many studies have borrowed attention alignment from the T2I model to improve multimodal capabilities. DIVA~\cite{wang2024diffusion} uses the image generation loss of the diffusion model to enhance the visual encoder of CLIP~\cite{radford2021learning}. LMFusion~\cite{shi2024llamafusion} empowers pretrained text-only LLMs with multimodal generative capabilities by sharing the self-attention layers with the diffusion model, which allows interactions across text and image features. The latest work, Diffusion Instruction Tuning~\cite{jin2025diffusion}, uses the text-image attention maps of the diffusion model as a guiding objective for the attention of the target VLM. Inspired by this, we reverse this process, using the grounding model's output as a guiding objective for the attention of the target T2I model, to improve the text-image alignment in the T2I model.

\section{Method}
To address word-level misalignment, we propose a Text-Image Alignment (TIA) framework, illustrated in Figure~\ref{fig:method} (a). Utilizing a grounding model, we obtain word-level segmentation masks for the input query. This allows for improved alignment between the word for text rendering and the corresponding attention map. To optimize the utilization of large Diffusion Transformer (DiT) based T2I models, we develop WordCon, a novel plug-and-play hybrid PEFT method, depicted in Figure~\ref{fig:method} (b). Unlike prior works~\cite{butt2025colorpeel,kumari2023multi} that directly fine-tune selected key parameters, WordCon reparameterizes the key parameters to enhance computational efficiency and usage flexibility. Finally, Figure~\ref{fig:method} (c) presents the plug-and-play inference pipeline. By integrating it with an image-conditioned pipeline, we showcase the practicality and portability of our proposed approach.

\begin{figure*}[t]
  \centering
  \includegraphics[width=\textwidth]{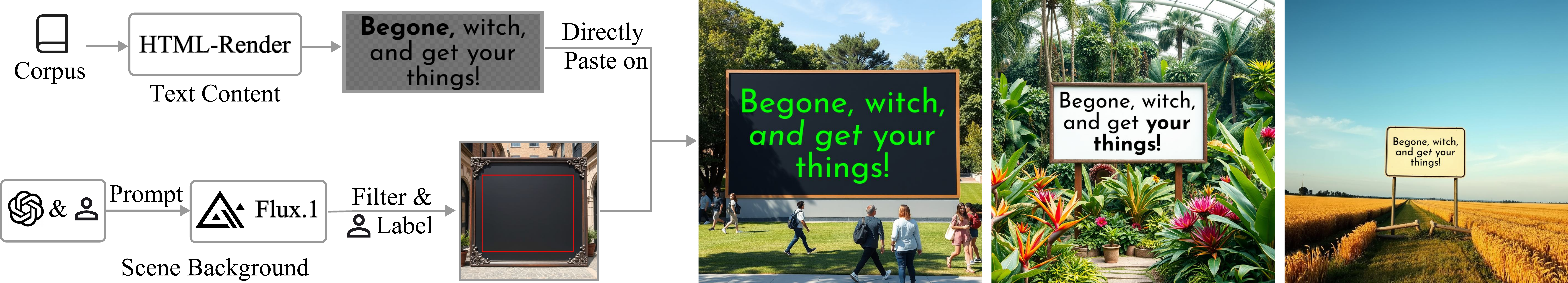}
  \vspace{-6mm}
  \caption{Our dataset construction pipeline. The regions of text are in different scales, and the text is in different scene backgrounds.}
  \vspace{-3mm}
  \label{fig:data}
\end{figure*}

\subsection{Preliminary}
\noindent \textbf{Preliminaries of Rectified Flow DiT.}
To avoid the computationally expensive process of the ordinary differential equation (ODE) $dy_t = v_\Theta(y_t, t)\,dt$, diffusion transformers such as \cite{esser2024scaling, blackforestlabs2024} directly regress a vector field $u_t$ that generates a probability path between noise distribution $p_1$ and data distribution $p_0$. To construct such a vector field $u_t$, \cite{esser2024scaling} considers a forward process that corresponds to a probability path $p_t$ transitioning from $p_0$ to $p_1=\mathcal{N}(0, 1)$. This can be represented as $z_t = a_t x_0 + b_t \epsilon\text{, where}\;\epsilon \sim \mathcal{N}(0,I)$.
With the conditions $a_0 = 1, b_0 = 0, a_1 = 0$ and $b_1 = 1$, the marginals $p_t(z_t) =
  \mathbb{E}_{\epsilon \sim \mathcal{N}(0,I)}
  p_t(z_t \vert \epsilon)\;$ align with data and noise distribution. Referring to \cite{lipman2023flow,esser2024scaling}, the marginal vector field $u_t$ can generate the marginal probability paths $p_t$, using the conditional vector fields as follows:
\begin{align}
    u_t(z) = \mathbb{E}_{\epsilon \sim
  \mathcal{N}(0,I)} u_t(z \vert \epsilon) \frac{p_t(z \vert
  \epsilon)}{p_t(z)},
  \label{eq:marginal_u}
\end{align}

The conditional flow matching objective is formulated as:
\begin{align}
   {\mathcal{L}_{CFM}} =  \mathbb{E}_{t, p_t(z | \epsilon), p(\epsilon) }|| v_{\Theta}(z, t) - u_t(z | \epsilon)  ||_2^2\;, 
   \label{eq:condflowmatch}
\end{align}
where the conditional vector fields $u_t(z \vert \epsilon)$ provides a tractable and equivalent objective. 

\subsection{Text-image alignment (TIA) Framework}
The TIA Framework presents a general approach for addressing the persistent challenge of misalignment in text-to-image generation models, including word-level misalignment in our case. As illustrated in Figure~\ref{fig:method} (a), TIA creates dual information flows between text-to-image (T2I) models and grounding models or vision-language models (VLMs). The framework starts with a shared input image sample, a prompt for the T2I model, and a query for the grounding model. The grounding model generates precise word-level segmentation masks, which act as spatial anchors for typography control. These masks are then utilized to guide the T2I model through two specialized loss functions: $\mathcal{L}_{Mask}$ and $\mathcal{L}_{Attn}$. The masked loss ($\mathcal{L}_{Mask}$) increases the weight of text regions in the image during the training process. In contrast, the attention loss ($\mathcal{L}_{Attn}$) explicitly aligns the model's attention maps with the segmentation masks. This process encourages the T2I model to establish more precise links between words and their exclusive regions in the image. The framework's intermediate results visualize masked latent representations, attention maps for each word (e.g., `JUMP', `OVER'), and the corresponding segmentation masks, which will be further utilized in WordCon.

\subsection{WordCon: A Plug-and-Play Hybrid PEFT Method}
\noindent \textbf{Attention of DiT.}
The backbone network of Flux contains two types of DiT layers: 19 Double-DiT layers and 38 Single-DiT layers. Inspired by ~\cite{helbling2025conceptattention, shi2024fonts, hu2025dcedit}, we interpret the joint attention mechanism of Double-DiT layers (combining text and image attention in Figure ~\ref{fig:method} (b)) as a form of `cross-attention' that aligns and fuses information from two different modalities. In contrast, the multimodal attention mechanism (green blocks in Figure ~\ref{fig:method} (b))in Single-DiT layers functions as a `self-attention' mechanism, primarily focusing on its own input.

\noindent \textbf{Selective Reparameterization.}
Following this perspective, the text attention within the joint attention mechanism of Double-DiT layers can be viewed as analogous to the key and value components (`k' and `v') in the cross-attention mechanism of the UNet, which are primarily responsible for incorporating text-based conditional control. This concept is consistent with approaches in~\cite{kumari2023multi, butt2025colorpeel, shi2024fonts}. However, unlike these methods, we opt to reparameterize the text attention rather than fine-tune it directly. This choice aims to reduce computational cost and enhance portability.

\noindent \textbf{Masked Loss.} 
To guide the model to focus on the text region in the image, we extend Eq. \ref{eq:condflowmatch} to highlight the textual regions in the current step. Specifically, within a prompt containing \textit{k} words which are desired to be generated, the $\mathcal{L}_{CFM}$ of the corresponding pixels is obtained from the latent mask of each word. Formally, considering $\textit{M}_{k}=\bigvee_{i=1}^{k}\textit{M}_{i}$ to the union of the pixels of \textit{k} words, the masked conditional flow matching loss is formulated as:

\begin{equation}
  {\mathcal{L}_{mask}} =  \mathbb{E}_{t, p_t(z | \epsilon), p(\epsilon)}|| \textit{M}_k(v_{\Theta}(z, t) - u_t(z | \epsilon))  ||_2^2\;, 
  \label{eq:masked_loss}
\end{equation}

\noindent \textbf{Joint Attention Loss.} 
The masked loss is designed to focus on all words in the text region of an image. Additionally, to ensure that each word for text rendering can align to its corresponding word-level position, we extend the cross-attention loss~\cite{avrahami2023break} from UNet into the joint-attention loss in the DiT model. This loss encourages words to pay attention exclusively to their corresponding target regions, which is formulated as:

\begin{equation}
    \mathcal{L}_{attn}=\mathbb{E}_{\textit{z}, c, t} \parallel J_{attn}(\textit{z}_t,\psi_{\vartheta}(c)_i) -\textit{M}_i \parallel_2^2.
    \label{eq:joint_attn_loss}
\end{equation}

Here, $J_{attn}(\textit{z}_t,\psi_{\vartheta}(c)_i)$ represents the joint-attention map between the visual representation $\textit{z}_{t}$ at the current step $t$ and the token $\psi_{\vartheta}(c)_i$ of the text condition $\textit{c}$, while $\textit{M}_{i}$ denotes the word-level mask of $\psi_{\vartheta}(c)_i$.  Thus, the total loss used is in the following, where the $\lambda_{attn} = 0.01$.

\vspace{-1mm}
\begin{equation}
  \mathcal{L}_{total}= \mathcal{L}_{mask} + \lambda_{attn}\mathcal{L}_{attn}.
  \label{eq:total_loss}
\end{equation}

\subsection{Dataset Construction} \label{sec:dataset}
Our dataset construction pipeline introduces a comprehensive approach for synthesizing text images in various scene contexts with controlled typographic attributes. As illustrated in Figure~\ref{fig:data}, the pipeline begins with a corpus that provides the foundation text content, which is processed through an HTML-Renderer to create the initial text image with a transparent background. At the same time, scene backgrounds are created using Flux.1 models guided by specific prompts, followed by quality filtering and labeling processes that ensure reasonable text placement and text size. Then, the rendered text images with transparent layers are directly pasted onto scene backgrounds, and the resulting dataset captures text in diverse scenarios with varying scales and backgrounds, enabling fine-grained typography control for specific words in the generated images. This carefully constructed dataset is essential for training models to recognize and manipulate typographic attributes such as bold, italic, or underlined styles selectively at the word level while maintaining visual coherence with the surrounding scene content.



\section{Experiments}
\subsection{Settings}
\textbf{Dataset.} 
We used the dataset constructed in Section~\ref{sec:dataset} for training. It comprises 28,000 samples featuring five different font types: serif, non-serif, sans-serif, script, and monospace. Text lengths range from 3 to 70 characters, with the majority falling between 20 and 50 characters. All images were resized to a resolution of 512 $\times$ 512 pixels to facilitate faster training. More details of the datasets are presented in Sec.H of the supplementary material.

\begin{table}[t]
  \centering
      \setlength{\tabcolsep}{3pt} 
       \begin{tabularx}{\columnwidth}{Xccccccc}
          \toprule
          \multirow{2}{*}{Methods} & \multicolumn{3}{c}{Controllability} & \multicolumn{2}{c}{Image Quality} & \multicolumn{2}{c}{OCR Accuracy} \\[0pt]
          \cmidrule(lr){2-4} \cmidrule(lr){5-6} \cmidrule(lr){7-8}
          ~ & \vspace{-2mm}\scriptsize Type${\uparrow}$ & \scriptsize Word${\uparrow}$ & \scriptsize Total${\uparrow}$ & \scriptsize Aesthetic${\uparrow}$ & \scriptsize Quality${\uparrow}$ & \scriptsize Prec.${\uparrow}$ & \scriptsize Recall${\uparrow}$ \\[-0.6mm]
          \midrule
          SD3 & 19.05 & 19.05 & 9.52 & 68.06 & 94.45 & 52.48 & 52.01 \\
          Flux & 23.81 & 14.29 & 14.29 & \underline{75.62} & \textbf{97.94} & 45.36 & 43.25 \\
          Recraft & 23.81 & 33.33 & 19.05 & 72.91 & 96.52 & 71.52 & 67.33 \\
          Gemini & 47.62 & 28.57 & 23.81 & 73.47 & 94.24 & 49.52 & 53.14 \\
          GPT4o & \underline{66.67} & 52.38 & 47.62 & 71.68 & 93.47 & \textbf{85.00} & \textbf{83.65} \\
          Ideogram & 61.90 & \underline{66.67} & \underline{52.38} & \textbf{76.21} & 94.45 & 69.90 & 65.48 \\
          \textbf{Ours} & \textbf{80.95} & \textbf{71.43} & \textbf{71.43} & 73.50 & \underline{96.55} & \underline{83.14} & \underline{81.95} \\
          \bottomrule 
      \end{tabularx}
  \caption{Quantitative comparison of our method with state-of-the-art models across word-level control accuracy (controllability), image quality, and OCR accuracy metrics. The bold and underlined numbers indicate the best and second-best performance in each column.}
  \vspace{-4mm}
  \label{tab:quant_compare}
  \vspace{-6mm}
\end{table}

\noindent \textbf{Implementation Details.} 
Experiments were conducted using the FLUX.1-dev model as the base. All runs were performed on a single H20 GPU with 96GB of memory for 90,000 training steps with a total batch size of 16. Following~\cite{xie2023omnicontrol}, we employed the Prodigy optimizer, and the default LoRA rank was set to 16.

\noindent \textbf{Metrics.} 
To evaluate the \textit{Controllability} of word-level typographic attributes, we followed~\cite{wang2025designdiffusion, wang2025beyond} and used GPT4o~\cite{openai_gpt4o} for recognizing typographic attributes in the generated images. For \textit{Image Quality}, we adopted the approach from~\cite{zhao2025lex}, utilizing Q-Align~\cite{wu2024q} (a large multimodal model dedicated to visual scoring) to assess both aesthetic and quality. The \textit{quality} primarily concerns the impact of distortions and other issues on human perception. In contrast, higher-level attributes such as content, lighting, color, and composition are considered more significant for image \textit{aesthetic} assessment~\cite{kong2016photo}. For \textit{OCR Accuracy}, we followed~\cite{tuo2024anytext,shi2024fonts} by using PaddleOCR~\cite{PaddleOCR} to calculate the precision and recall of OCR results. All these metrics are reported as percentages, though the percentage sign has been omitted in the tables for aesthetic reasons.

\subsection{Comparisons}
We selected the same T2I models used in~\cite{chen2025posta} for comparisons, which includes general models such as SD3~\cite{esser2024scaling}, FLUX-dev~\cite{blackforestlabs2024}, GPT4o-img~\cite{gpt4oimg}, and Gemini2~\cite{gimini2024}, and design-oriented models known for their strong text rendering capabilities, namely Recraft v3~\cite{recraft2024} and Ideogram v3~\cite{ideogram2024}.

\begin{figure*}[t]
  \centering
  \includegraphics[width=\textwidth]{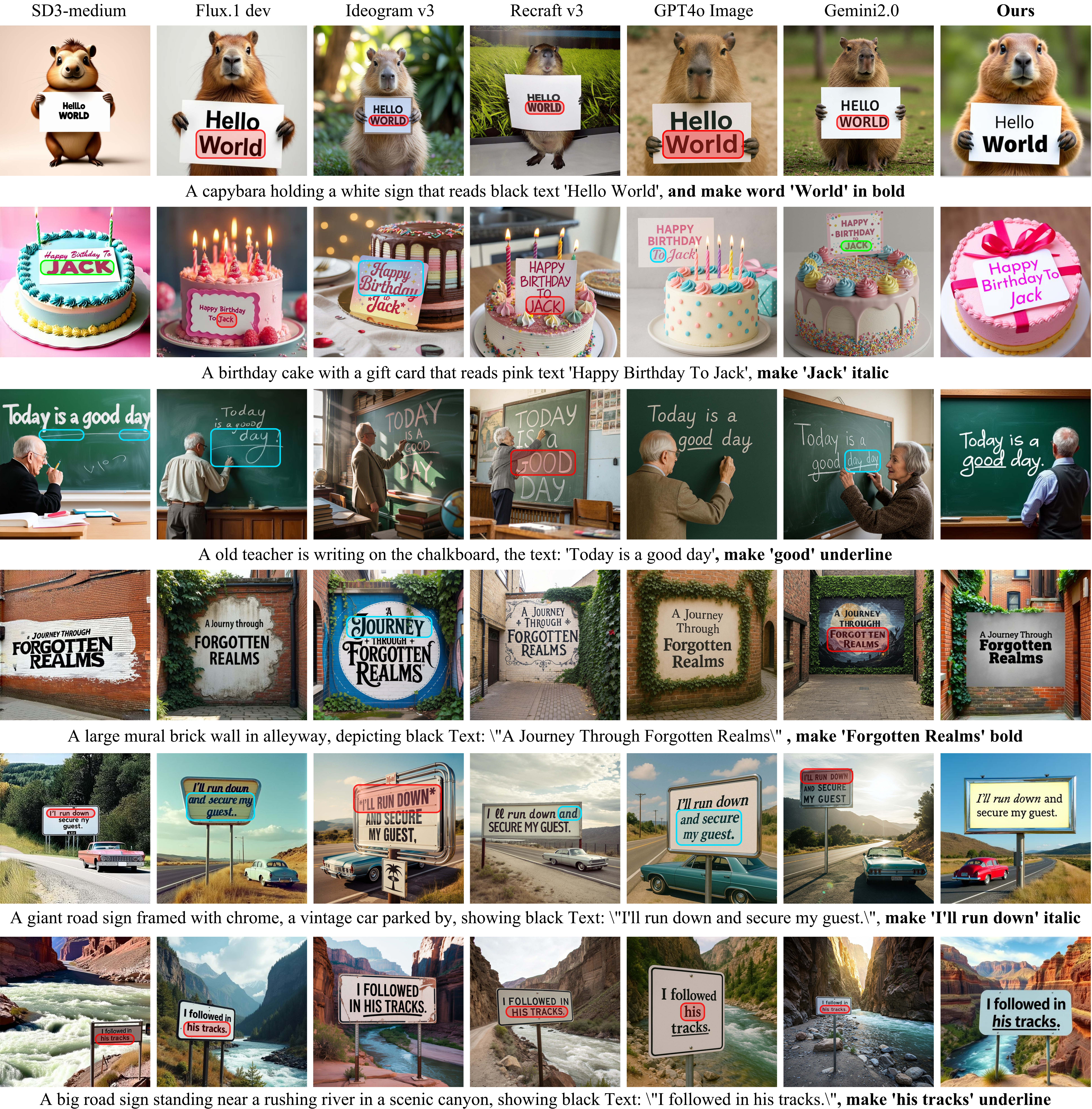}
  \caption{Qualitative comparison with state-of-the-art models, including both widely-used open-source and proprietary commercial models. The first three rows of the comparison illustrate cases where typography control is applied to a single word, while the last three rows demonstrate control over multiple words. \textcolor{red}{Red boxes} indicate instances where no typographic attribute was applied to the target word. \textcolor[HTML]{00DDFF}{Blue boxes} denote cases where the correct typographic attribute was applied to an incorrect word. \textcolor[HTML]{00FF00}{Green boxes} highlight instances where an incorrect typographic attribute was applied to the target word.}
  \label{fig:compare}
\end{figure*}

\begin{figure*}[t]
  \centering
  \includegraphics[width=\textwidth]{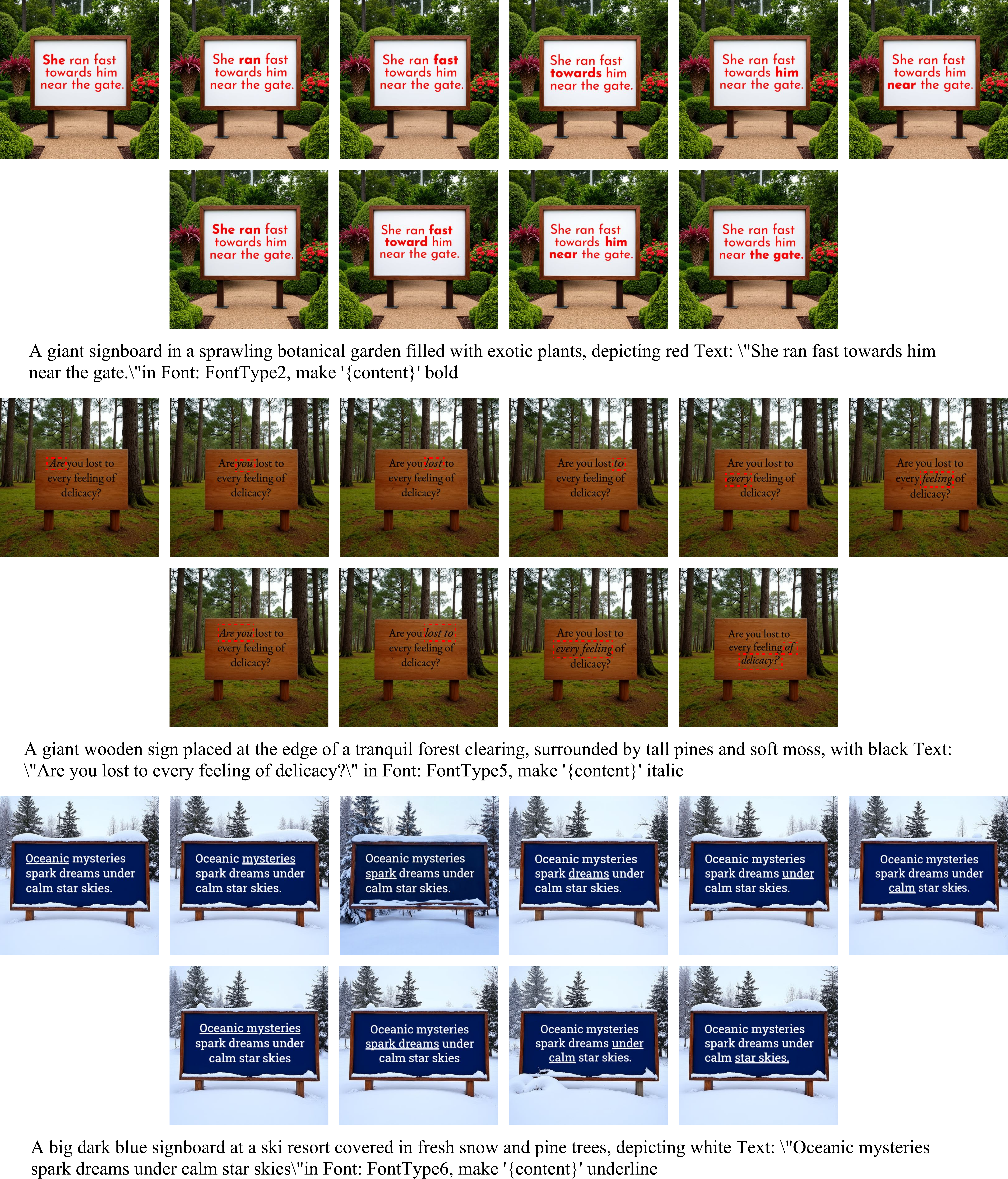}
  \caption{More results of our method. The cases in the first, third, and fifth rows where typography control is applied to a single word, while the other three rows demonstrate control over multiple words. The results show that our method can achieve word-level typography control while also enabling font control.}
  \label{fig:more_results}
\end{figure*}

\noindent \textbf{Qualitative Results.}
To evaluate the effectiveness of our method, we conducted a comprehensive qualitative comparison with these models as shown in Figure~\ref{fig:compare}, and Figure~\ref{fig:more_results} presents more qualitative results of our method. As illustrated, ours significantly outperforms the others on accurate word-level controllability, and at the same time, the aesthetic quality is comparable to that of the top proprietary commercial models. For the three failure cases (red, blue, green boxes), the most common one for existing models is the red box case, where no typographic attribute was applied to the target word. Followed by a correct typographic attribute on an incorrect word (blue boxes) and an incorrect typographic attribute on the correct target word (green boxes). In comparison, Ideogram and GPT4o-img achieved the highest number of completely correct results in text rendering, consistent with our quantitative findings. Additionally, we noticed that Recraft, Ideogram, and Gemini had particular difficulty matching uppercase and lowercase letters in the prompt, with generated images tending to be all uppercase and not respecting the case differences in the original prompt.

\noindent \textbf{Quantitative Results.}
We also conducted a quantitative analysis comparing our method with these six models through 21 prompts, which contained different typographic attributes evenly. The results are shown in Table~\ref{tab:quant_compare}. We can observe the following quantitative analysis results. In terms of controllability, our proposed method achieved the best performance across all three metrics: Type control at 80.95\%, Word control at 71.43\%, and Total control at 71.43\%, significantly outperforming other comparative models. Compared to the second-best models, our method shows improvements of 14.28 percentage points in Type control, 4.76 percentage points in Word control, and 19.05 percentage points in Total control. Regarding image quality, although our method did not achieve the highest scores, it received an Aesthetic score of 73.50 and a Quality score of 96.55, ranking third and second, respectively, very close to the best performances. For OCR accuracy, our method performed excellently with a Precision of 83.14\% and Recall of 81.95\%, ranking second only to GPT4o-img. Compared to other models, our method demonstrates a significant advantage regarding text accuracy. Overall, the proposed method exhibits a substantial lead in controllability while maintaining high image quality and achieving near-optimal text accuracy, indicating its comprehensive and balanced performance in word-level controlled scene text rendering.

\begin{figure}[t]
  \centering
  \includegraphics[width=\columnwidth]{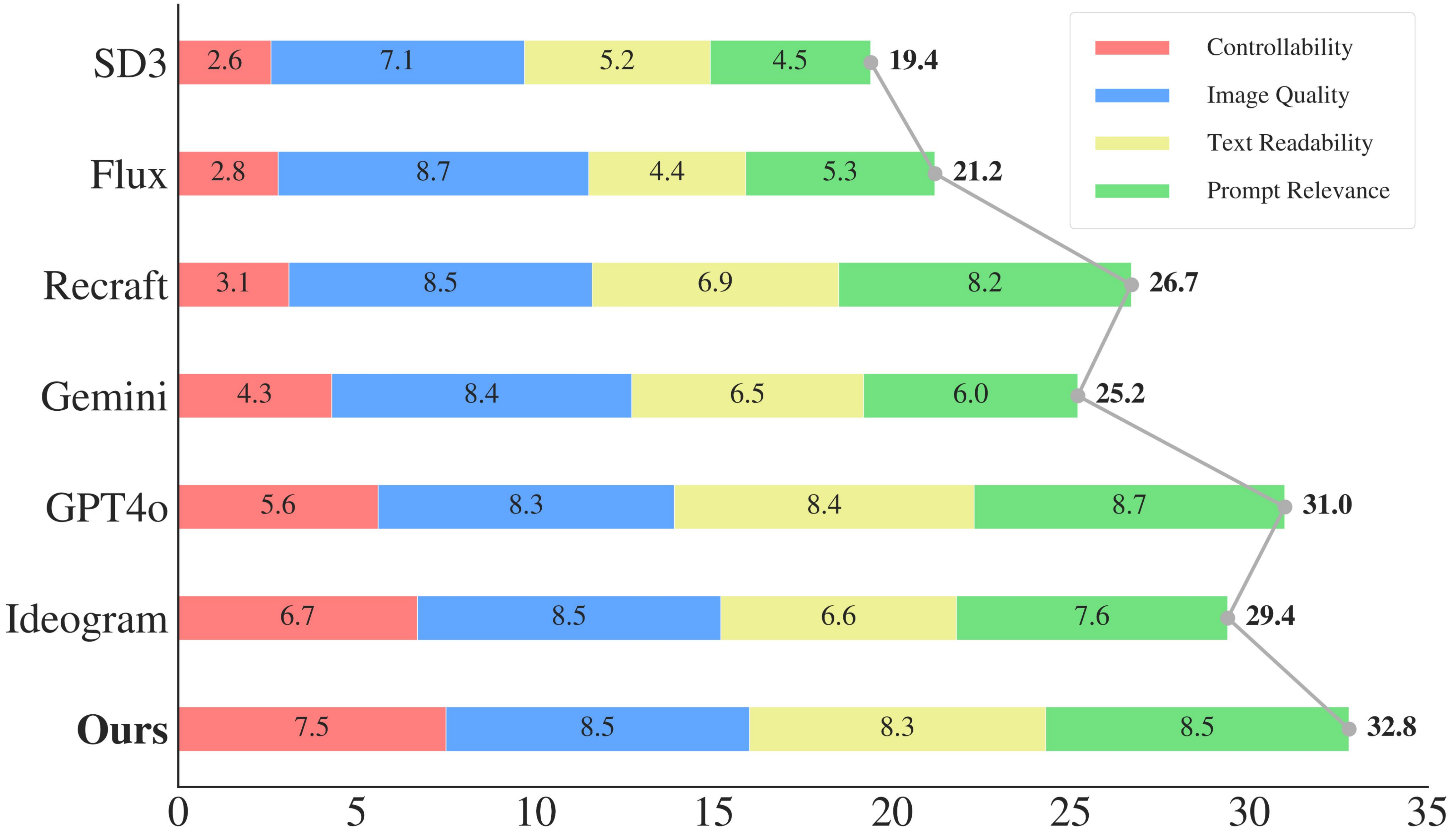}
  \vspace{-6mm}
  \caption{Human evaluation results}
  \vspace{-6mm}
  \label{fig:user_study}
\end{figure}

\noindent \textbf{User Study.}
In addition to the automated evaluation, we invited twenty users to conduct the human evaluation. These users with experience in AI tools and a design background assess key aspects such as controllability at word-level typographic attributes, image quality, text readability, and prompt-image relevance, the rating ranges from 1 to 10. As shown in Figure~\ref{fig:user_study}, our method achieves the highest overall score (32.8), outperforming all baselines. Notably, it achieves the top score in controllability (7.5), significantly surpassing other models. It also ranks highly in image quality (8.5), text readability (8.3), and prompt relevance (8.5). In comparison, the commercial models GPT4o-img (31.0) and Ideogram (29.4) follow as the second and third-best performers, respectively. The open-sourced models SD3 and Flux exhibit notable shortcomings across multiple aspects. These results further validate that our approach not only enables word-level controllability but also preserves high-generation quality, text accuracy, and prompt-image alignment.

\subsection{Ablation Study}

\noindent \textbf{Ablation on loss.} 
To better understand the contribution of each component in our method, we randomly select 300 samples as the test set to conduct the ablation study, as shown in Table~\ref{tab:ablation}. The original setting means without fine-tuning, which is the original Flux-dev. The $\mathcal{L}_{Vanilla}$ means fine-tuning the base model with the vanilla loss in Eq.~\ref{eq:condflowmatch}. With the incorporation of the two losses, controllability has been notably enhanced. Additionally, the full model (last row) closely approaches the optimal values in terms of image quality and OCR accuracy. Besides, we also provide additional ablation on selected parameters and LoRA rank in Sec.E of the supplementary.

\begin{table}[t]
    \centering
    \setlength{\tabcolsep}{3pt} 
    \begin{tabular}{cccccccc}
      \toprule
        \multirow{2}{*}{Settings} & \multicolumn{3}{c}{Controllability} & \multicolumn{2}{c}{Image Quality} & \multicolumn{2}{c}{OCR Accuracy} \\
        \cmidrule(lr){2-4} \cmidrule(lr){5-6} \cmidrule(lr){7-8}
        ~  & \scriptsize Type${\uparrow}$ & \scriptsize Word${\uparrow}$ & \scriptsize Total${\uparrow}$ & \scriptsize Aesthetic${\uparrow}$ & \scriptsize Quality${\uparrow}$ & \scriptsize Prec.${\uparrow}$ & \scriptsize Recall${\uparrow}$ \\
        \midrule
        original & 23.00 & 20.00 & 8.67 & \textbf{72.36} & 96.41 & 65.87 & 63.70 \\
        $\mathcal{L}_{Vanilla}$ & 49.00 & 36.33 & 27.22 & 71.05 & \textbf{96.62} & \textbf{76.54} & \textbf{75.64} \\
        +$\mathcal{L}_{Mask}$ & \underline{58.00} & \underline{48.33} & \underline{42.67} & 69.42 & 95.60 & 71.91 & 74.67 \\
        +$\mathcal{L}_{Attn}$ & \textbf{77.67} & \textbf{68.00} & \textbf{66.67} & \underline{71.26} & \underline{95.88} & \underline{74.36} & \underline{75.34} \\
        \bottomrule
    \end{tabular}
    \caption{Ablation study on each component in our method. $\mathcal{L}_{Vanilla}$: vanilla loss. $\mathcal{L}_{Mask}$: masked loss. $\mathcal{L}_{Attn}$: joint attention loss.}
    \label{tab:ablation}
    \vspace{-6mm}
\end{table}

\subsection{Applications}

\begin{figure}[t]
  \centering
  \includegraphics[width=\columnwidth]{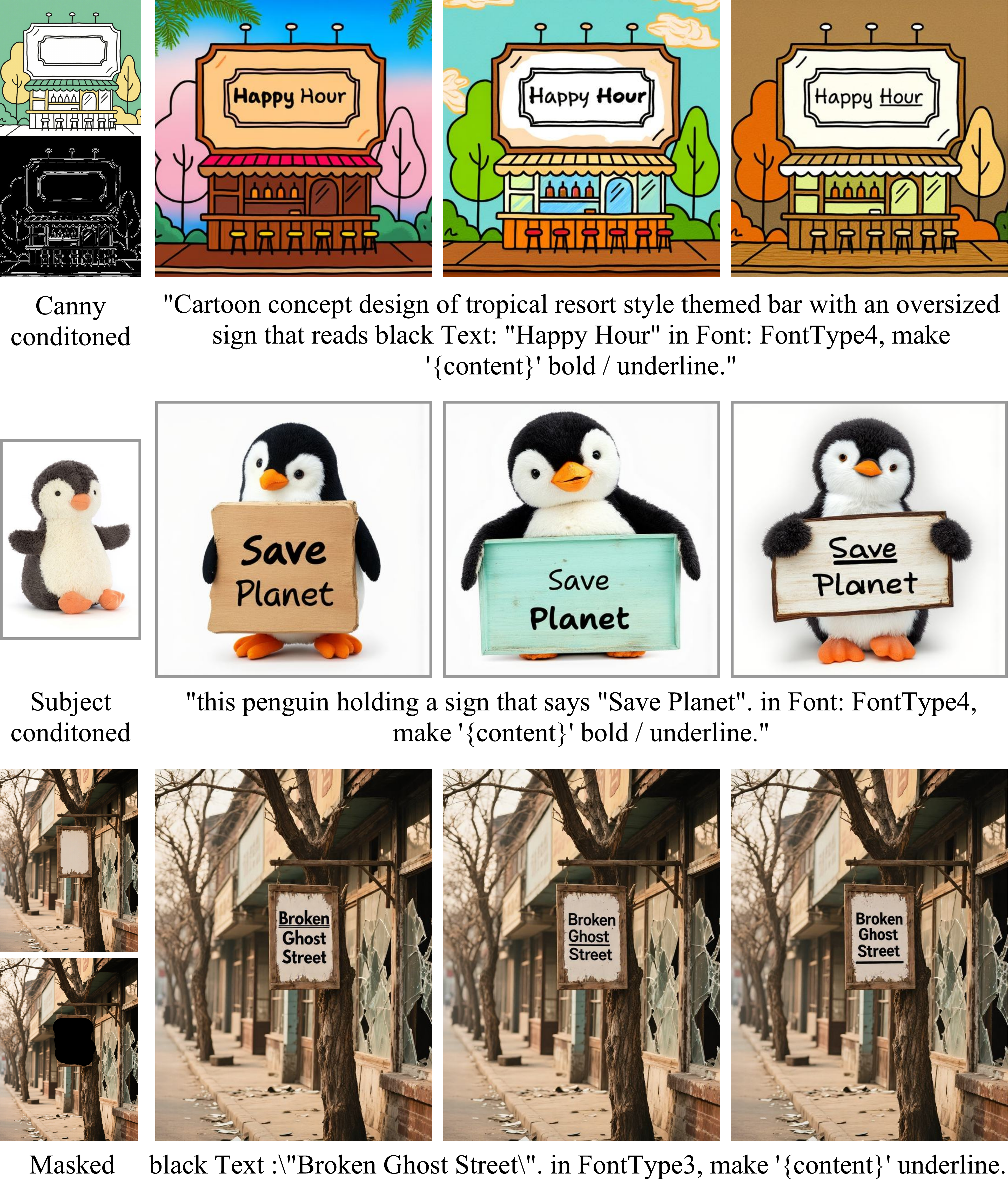}
  \vspace{-7mm}
  \caption{Visual results of various applications.}
  \vspace{-4mm}
  \label{fig:app}
\end{figure}

\begin{figure}[t]
  \centering
  \includegraphics[width=\columnwidth]{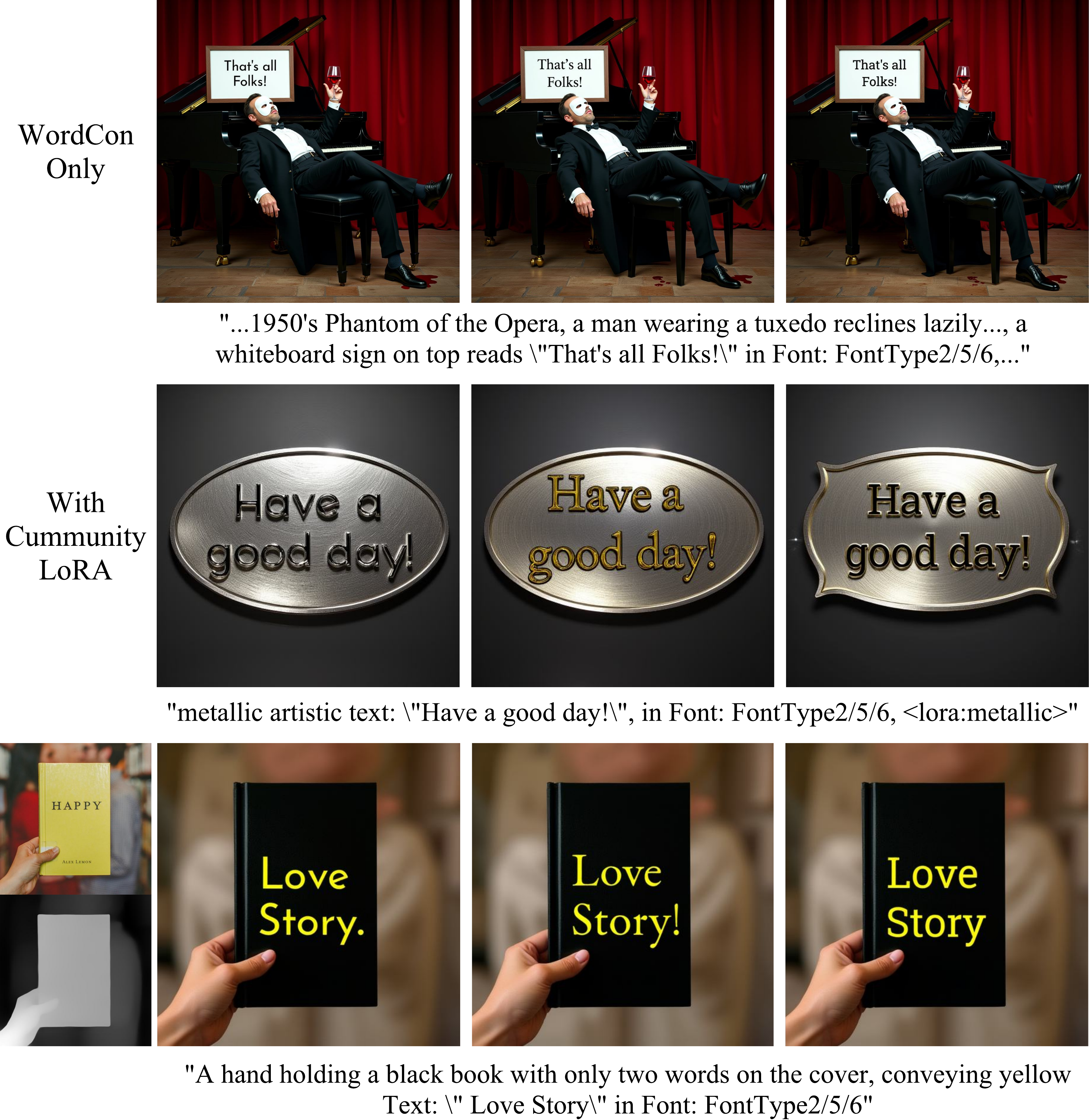}
  \vspace{-7mm}
  \caption{Visual results of font selection.}
  \vspace{-3mm}
  \label{fig:font_select}
\end{figure}

\noindent \textbf{Image-Conditioned Pipeline.}
As our word-level control is implemented via a text prompt, WordCon is compatible with image-conditioned pipelines, such as OminiControl~\footnote{\href{https://github.com/Yuanshi9815/OminiControl/tree/main}{OminiControl}}. Consequently, WordCon can support canny-conditioned (first row of Figure~\ref{fig:app}), subject-conditioned (second row of Figure~\ref{fig:app}), depth-conditioned (third row of Figure~\ref{fig:font_select}) word-level controllable text rendering, etc.

\noindent \textbf{Text Editing and Placement Control.} 
With Flux-fill~\footnote{\href{https://huggingface.co/black-forest-labs/FLUX.1-Fill-dev}{FLUX.1-Fill-dev}}, our method not only supports the regeneration of erased text (text editing, third row of Figure~\ref{fig:teaser}(b)), but also supports the generation of text within specified blank regions (placement control, third row of Figure~\ref{fig:app}). Additionally, while the WordCon is trained on 512 $\times$ 512 images, the integrated pipeline supports non-square output resolutions.

\noindent \textbf{Artistic Text Rendering.} 
WordCon is designed to control image content and can thus work with stylized (artistic font) LoRAs~\footnote{\href{https://civitai.com/models/662990?modelVersionId=741972}{ballon LoRA}} in the community, as demonstrated in the first row of Figure~\ref{fig:teaser}(b).

\noindent \textbf{Font Selection.} 
WordCon enables font selection in scene text rendering (first row in Figure~\ref{fig:font_select}). This capability is compatible with artistic LoRAs~\footnote{\href{https://civitai.com/models/531692?modelVersionId=772512}{metallic LoRA}} (second row in Figure~\ref{fig:font_select}) and image-conditioned pipelines (third row in Figure~\ref{fig:font_select}).

\subsection{Limitations}
We observe that when the prompt includes the same word multiple times, the results tend to be unstable, making it difficult to control a specific single instance accurately. Instead, the model tends to apply control to all instances. e.g., `Toward him I am kinder than toward myself'. The second `toward' will be affected somewhat when specifically controlling the first 'toward'. We discuss the limitations in Sec.G of the supplementary material.

\section{Conclusions}
We focus on word-level typography control in scene text rendering. To tackle word-level misalignment, we introduce the Text-Image Alignment (TIA) framework, harnessing the fine-grained text-image alignment of grounding models as supervision. As the vanilla loss falls short for word-level misalignment, we incorporate masked and attention losses to boost controllability. Given the increasing scale of T2I models, we propose WordCon, a plug-and-play hybrid PEFT approach that saves computational costs and enables integration into various applications. To facilitate training, we introduce a new word-level controlled scene text dataset and will continuously scale it up to cover more fonts and complex cases in further work.

\begin{acks}

The work described in this paper was substantially supported by a grant from the Research Grants Council of the Hong Kong Special Administrative Region, China (Project No. PolyU/RGC Project PolyU 25211424) and partially supported by a grant from PolyU university start-up fund (Project No. P0047675).

\end{acks}

\clearpage

\clearpage

\clearpage



\bibliographystyle{ACM-Reference-Format}
\bibliography{sample-bibliography}

\clearpage
\appendix

\end{document}